\documentclass[10pt,twocolumn,letterpaper]{article}

\usepackage[margin=0.75in]{geometry}
\usepackage[T1]{fontenc}
\usepackage{times}
\usepackage[numbers,sort&compress]{natbib}
\usepackage{amssymb}
\usepackage{xcolor}
\usepackage{tikz}
\usetikzlibrary{arrows.meta}
\definecolor{wacvblue}{rgb}{0.21,0.49,0.74}
\usepackage[pagebackref,breaklinks,colorlinks,allcolors=wacvblue]{hyperref}
\usepackage{booktabs}
\usepackage{multirow}
\usepackage{amsmath}
\usepackage{graphicx}

\title{Continuous Foundation-Guided Pseudo-Label Weighting for Breast Ultrasound Domain Adaptation}
\author{Xinyu Fan\thanks{These authors contributed equally.} \and Xinyi Li\footnotemark[1]}
\date{}

\begin{document}
\maketitle
\begin{abstract}
Breast ultrasound segmentation models often generalize poorly across scanners, institutions, and patient populations. We investigate \emph{continuous foundation-guided pseudo-label weighting} for unsupervised domain adaptation (UDA). An exponential-moving-average (EMA) teacher generates target-domain pseudo-labels, while a frozen ultrasound segmentation model (AutoSAMUS) modulates their contribution using a bounded weight that combines the prior's certainty with its probabilistic agreement with the teacher. In contrast to our earlier selective BiRP rule, the proposed formulation supervises every target pixel with a weight in $[0.5,1]$ and requires no foundation-model inference at test time. We evaluate the method on five public breast ultrasound datasets, spanning three source domains and eleven source--target directions. Under the TransUNet protocol, continuous weighting outperforms both all-pixel EMA and original BiRP on all four BUSI-source targets, and achieves the highest mean Dice and lowest HD95 among the compared methods on all four BUS-BRA-source targets. Results with BrEaST as the source are mixed: continuous weighting performs best on BUSI and comparably to original BiRP on OASBUD, whereas none of the adaptation methods surpasses source-only training on BUS-UCLM. In a separate U-Net pipeline, agreement-only weighting performs comparably to the full formulation, leaving the incremental value of prior certainty unresolved. Because the weighting rule was developed after some target-test results had been examined, we treat the study as exploratory and report adverse as well as favorable outcomes.
\end{abstract}

\section{Introduction}
\label{sec:intro}
Breast ultrasound is widely used for lesion detection and characterization because it is affordable, portable, and free of ionizing radiation. Automated lesion segmentation can facilitate quantitative measurement and downstream analysis; however, ultrasound appearance varies substantially across devices, transducers, acquisition settings, operators, and patient populations. Consequently, models trained on one public dataset often generalize poorly to another \cite{aldhabyani2020busi,gomezflores2024busbra,pawlowska2024breast,vallez2025bus}. Because acquiring pixel-level annotations at every new site is expensive, unsupervised domain adaptation (UDA) seeks to transfer knowledge from a labeled source domain to an unlabeled target domain.

Self-training is a practical and widely adopted strategy for segmentation UDA \cite{zou2018cbst,tranheden2021dacs,hoyer2022daformer}. Typically, a teacher model---often an exponential-moving-average (EMA) copy of the student \cite{tarvainen2017mean}---predicts pseudo-labels for target images, which are then used to train the student. A central limitation is confirmation bias: errors are reinforced when the teacher is uncertain or systematically affected by domain shift. Confidence thresholding can suppress unreliable supervision, but may also remove ambiguous boundaries and low-contrast regions---precisely the structures that adaptation should improve in ultrasound images.

Foundation segmentation models can provide an external supervisory signal that is independent of the student's optimization trajectory \cite{kirillov2023sam,ma2024medsam,lin2023autosamus}. Our earlier BiRP formulation combined a frozen ultrasound prior with an EMA teacher to \emph{select} reliable foreground and background pixels through bidirectional gates. Controlled comparisons, however, showed that this selective strategy did not consistently outperform all-pixel EMA self-training or direct use of the prior. Moreover, hard selection discards potentially useful supervision and introduces multiple thresholds whose appropriate values may vary across domains.

These observations raise a different question: can a fixed prior provide an informative \emph{weighting} signal without discarding target pixels or replacing the EMA teacher's labels? We investigate a continuous agreement rule that combines the frozen prior's certainty with its probabilistic agreement with the EMA teacher, producing a bounded weight in $[0.5,1]$. Every target pixel contributes to training, while pixels for which the prior is decisive and agrees with the teacher receive up to twice the weight assigned to the least trusted pixels. The teacher remains the sole source of pseudo-labels, and the prior is used only during training.

Our contributions are:
\begin{itemize}
\item We formulate a bounded per-pixel weighting rule that incorporates a frozen ultrasound prior into EMA self-training without hard pixel selection or additional inference-time cost (Sec.~\ref{sec:method}).
\item We evaluate the rule on five public breast ultrasound datasets, covering three source domains and eleven adaptation directions, and compare it with all-pixel EMA and original BiRP using three seeds per setting. We also report source-only baselines for the BUS-BRA- and BrEaST-source directions (Sec.~\ref{sec:experiments}).
\item We provide component, pipeline, and qualitative analyses, report directions in which adaptation fails to help, and explicitly delimit the strength of the resulting evidence (Sec.~\ref{sec:limitations}).
\end{itemize}
Because the method was refined after some target-test outcomes had been inspected, we regard all reported results as exploratory rather than independently confirmatory.

\section{Related Work}
\label{sec:related}
\paragraph{Domain-adaptive segmentation.}
Approaches to UDA for semantic segmentation include input-space translation and Fourier-domain transfer \cite{yang2020fda}, feature- or output-level adversarial alignment \cite{ganin2016dann,vu2019advent}, and joint image--feature alignment for cross-modality medical imaging \cite{chen2019sifa}. Self-training with pseudo-labels provides a widely used alternative, with representative methods based on class-balanced self-training \cite{zou2018cbst}, cross-domain mixed sampling \cite{tranheden2021dacs}, prototypical pseudo-label denoising \cite{zhang2021proda}, and transformer architectures with multiresolution training \cite{hoyer2022daformer,hoyer2022hrda}. Many of these methods build on Mean Teacher \cite{tarvainen2017mean}, in which an EMA model provides more stable training targets. Medical-imaging studies have further addressed cross-site and cross-modality shifts \cite{zhang2024mapseg,stenger2025resampl}. Whereas most pseudo-labeling methods estimate reliability from the teacher's own confidence, our formulation derives an auxiliary weight from an independent frozen model.

\paragraph{Foundation models as priors.}
Segment Anything (SAM) \cite{kirillov2023sam} and medical variants such as MedSAM \cite{ma2024medsam} offer broadly transferable segmentation representations, while AutoSAMUS \cite{lin2023autosamus} adapts SAM to ultrasound through automatic prompting. A growing body of work uses SAM-family models to improve supervision under limited annotation. For example, SAM-Match \cite{xu2025sammatch} combines a teacher--student framework with a SAM-based model for semi-supervised medical segmentation, using teacher predictions as prompts for label refinement by a fine-tuned SAM component. Related methods employ foundation models to generate or correct pseudo-labels, or exploit foundation-model features and prompts to guide adaptation \cite{zhang2024iplc,zhang2025iplcplus,tejero2025samda,liu2025harp,zhang2025dpgs,wei2025nora}. In contrast, our approach keeps the prior frozen and uses it only to \emph{weight} labels generated by the EMA teacher. Consequently, an erroneous prior can reduce a pixel's weight by at most half but cannot reverse its pseudo-label.

\paragraph{Breast ultrasound segmentation.}
Breast ultrasound lesion segmentation is commonly addressed using encoder--decoder networks \cite{long2015fcn,ronneberger2015unet,chen2017deeplab} or hybrid CNN--transformer architectures such as TransUNet \cite{chen2021transunet}. Public datasets now encompass multiple countries and imaging systems, including BUSI \cite{aldhabyani2020busi}, BUS-BRA \cite{gomezflores2024busbra}, BrEaST \cite{pawlowska2024breast}, OASBUD \cite{piotrzkowska2017oasbud}, and BUS-UCLM \cite{vallez2025bus}. Differences in scanners, fields of view, annotation protocols, and lesion composition make these datasets a useful testbed for cross-dataset adaptation. Foundation models, EMA teachers, and agreement-based weighting are not individually novel; the contribution studied here is their specific bounded combination and its evaluation across a broad set of breast-ultrasound domain shifts.

\section{Method}
\label{sec:method}
\paragraph{Problem and model.}
Let $\mathcal D_s=\{(x_s,y_s)\}$ denote the labeled source set and
$\mathcal D_t=\{x_t\}$ the unlabeled target set. In the main experiments, a TransUNet student
$f_\theta$ \cite{chen2021transunet} is trained with source labels and target
pseudo-labels. Its EMA teacher $f_{\bar\theta}$ is updated after each student
optimization step as $\bar\theta\leftarrow\alpha\bar\theta+(1-\alpha)\theta$.
The ultrasound prior $g$ is a frozen AutoSAMUS model
\cite{lin2023autosamus}. No gradients are propagated through either model
when constructing the pseudo-supervision.

\paragraph{Continuous agreement weight.}
For target pixel $u$, let $p(u)$ and $q(u)$ denote the foreground probabilities
predicted by the EMA teacher and the frozen prior, respectively. The teacher
provides the pseudo-label, $\hat y(u)=\mathbf 1[p(u)\ge 0.5]$. We define
prior certainty as $c(u)=2|q(u)-0.5|$ and probabilistic agreement as
$a(u)=1-|p(u)-q(u)|$, and then compute
\begin{equation}
 w(u)=0.5+0.5\,c(u)a(u). \label{eq:weight}
\end{equation}
Because $p,q\in[0,1]$, the resulting weight is bounded within $[0.5,1]$.
Every target pixel therefore contributes to training, while a decisive prior
that agrees with the teacher receives greater relative emphasis. Agreement is
not assumed to imply label correctness. Unlike original BiRP, the formulation
uses neither top-$r$ selection nor class-specific foreground/background gates.

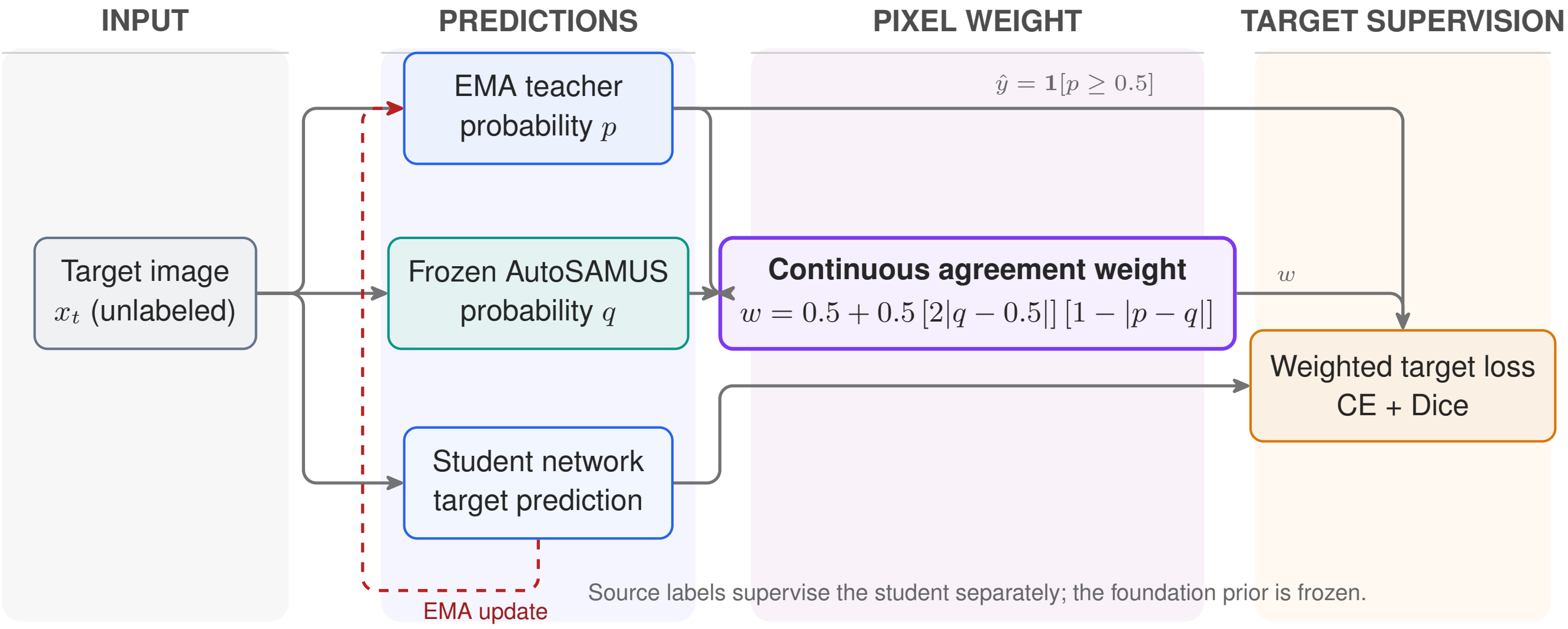
\begin{figure*}[!t]
\centering
\resizebox{\textwidth}{!}{\begin{tikzpicture}[
  font=\sffamily,
  >={Stealth[round,length=2.4mm,width=2.0mm]},
  panel/.style={draw=black!55, line width=.75pt, rounded corners=4pt,
    align=center, minimum height=12mm, inner sep=6pt, text=black!85,
    font=\sffamily\small},
  flow/.style={-> , draw=black!55, line width=.95pt, rounded corners=4pt},
  updatearrow/.style={-> , draw=updateclr!85!black, line width=.95pt, dashed, rounded corners=4pt},
  lane/.style={rounded corners=6pt, draw=none},
  collabel/.style={font=\sffamily\bfseries\small, text=black!70},
  smalllabel/.style={font=\sffamily\scriptsize, text=black!60, align=center}]

\definecolor{inputclr}{HTML}{64748B}
\definecolor{teacherclr}{HTML}{2563EB}
\definecolor{priorclr}{HTML}{0D9488}
\definecolor{studentclr}{HTML}{2563EB}
\definecolor{weightclr}{HTML}{7C3AED}
\definecolor{lossclr}{HTML}{D97706}
\definecolor{updateclr}{HTML}{DC2626}

\fill[lane, fill=black!3]    (-1.55,-3.55) rectangle ( 1.55, 2.65);
\fill[lane, fill=blue!4]     ( 2.55,-3.55) rectangle ( 5.95, 2.65);
\fill[lane, fill=violet!5]   ( 6.55,-3.55) rectangle (11.45, 2.65);
\fill[lane, fill=orange!5]   (12.0,-3.55) rectangle (15.2, 2.65);

\node[collabel] at (0,2.95)    {INPUT};
\node[collabel] at (4.25,2.95) {PREDICTIONS};
\node[collabel] at (9,2.95)    {PIXEL WEIGHT};
\node[collabel] at (13.6,2.95) {TARGET SUPERVISION};
\draw[draw=black!20, line width=.5pt] (-1.55,2.6) -- (1.55,2.6);
\draw[draw=black!20, line width=.5pt] (2.55,2.6) -- (5.95,2.6);
\draw[draw=black!20, line width=.5pt] (6.55,2.6) -- (11.45,2.6);
\draw[draw=black!20, line width=.5pt] (12.0,2.6) -- (15.2,2.6);

\node[panel, draw=inputclr, fill=inputclr!10, minimum width=2.3cm]
  (target) at (0,0) {Target image\\[1pt]$x_t$ (unlabeled)};
\node[panel, draw=teacherclr, fill=teacherclr!9, minimum width=2.9cm]
  (teacher) at (4.25,2) {EMA teacher\\[1pt]probability $p$};
\node[panel, draw=priorclr, fill=priorclr!11, minimum width=2.9cm]
  (prior) at (4.25,0) {Frozen AutoSAMUS\\[1pt]probability $q$};
\node[panel, draw=studentclr, fill=studentclr!6, minimum width=2.9cm]
  (student) at (4.25,-2.05) {Student network\\[1pt]target prediction};
\node[panel, draw=weightclr, line width=1.15pt, fill=weightclr!8, minimum width=4.5cm]
  (weight) at (9,0) {\textbf{Continuous agreement weight}\\[2pt]$w=0.5+0.5\,[2|q-0.5|]\,[1-|p-q|]$};
\node[panel, draw=lossclr, fill=lossclr!11, minimum width=3.2cm]
  (loss) at (13.6,-1) {Weighted target loss\\[1pt]CE + Dice};

\draw[flow] (target.east) -- ++(.5,0) |- (teacher.west);
\draw[flow] (target) -- (prior);
\draw[flow] (target.east) -- ++(.5,0) |- (student.west);
\draw[flow] (teacher.east) -- ++(.4,0) |- (weight.west);
\draw[flow] (prior) -- (weight);
\draw[flow] (weight.east) -| node[pos=.15, above, smalllabel] {$w$} (loss.north);
\draw[flow] (student.east) -- ++(.5,0) |- (loss.west);
\draw[flow] (teacher.east) -- ++(.8,0) -- node[above, smalllabel] {$\hat y=\mathbf{1}[p\geq0.5]$} (13.6,2) -- (loss.north);
\draw[updatearrow] (student.south) -- ++(0,-.55) -| node[pos=.15, below, smalllabel, text=updateclr!70!black] {EMA update} (2.35,2) -- (teacher.west);

\node[smalllabel] at (9,-3.25) {Source labels supervise the student separately; the foundation prior is frozen.};
\end{tikzpicture}}
\caption{Continuous foundation-guided target supervision. The EMA teacher
generates pseudo-labels, and its foreground probability is combined with that
of the frozen prior to determine the strictly positive weight in
Eq.~\eqref{eq:weight}. Source labels provide a separate supervised signal. In
the main TransUNet protocol, evaluation uses the EMA parameters associated
with the checkpoint selected on source-validation performance.}
\label{fig:method}
\end{figure*}

\paragraph{Training objective.}
The source objective $\mathcal L_s$ combines cross-entropy and soft Dice
losses. Following a source-only warm-up, target supervision comprises weighted
cross-entropy and weighted foreground Dice losses:
\begin{equation}
\mathcal L=\mathcal L_s+\lambda_p(\mathcal L_{\rm CE}^{w}+\mathcal L_{\rm Dice}^{w}).
\end{equation}
The cross-entropy term is normalized by the sum of the pixel weights. For the
Dice term, the same weights are applied to both the intersection and the
denominator on a per-image basis. The prior is used only during training; at
test time, the selected checkpoint requires a single segmentation-network
forward pass and no AutoSAMUS inference.

\paragraph{Properties of the weight.}
The product $c(u)a(u)$ is large only when the prior is both decisive and close
to the teacher. If the prior is uncertain ($q\approx0.5$), then $w\approx0.5$
irrespective of the teacher prediction, and the pixel receives the minimum unnormalized weight. If the prior is decisive but conflicts with the teacher,
$a(u)$ is small and the weight again approaches $0.5$. Thus, the prior never
changes the pseudo-label; it only rescales its contribution to the loss. The
positive lower bound retains a contribution from every pixel; the normalized
relative contribution also depends on weights elsewhere in the loss. Apart
from binarizing $p$ at $0.5$, the rule introduces no thresholds or learnable
parameters.

\begin{figure}[!htbp]\centering
\includegraphics[width=\linewidth]{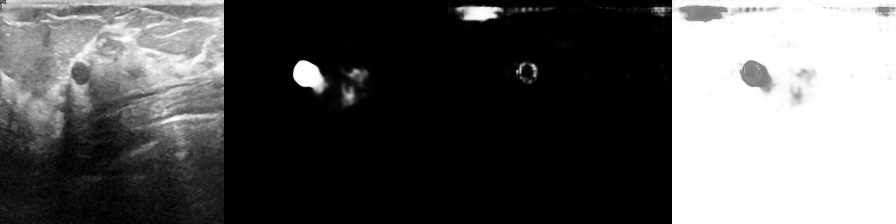}
\caption{Illustrative BrEaST$\rightarrow$BUSI test case: input, EMA-teacher probability, frozen-prior probability and continuous weight, from left to right. This selected case illustrates the weight map but is not representative of outcome frequency.}
\label{fig:weightmap}\end{figure}

Figure~\ref{fig:weightmap} illustrates the weight on a held-out target image.

\paragraph{Compared target-supervision rules.}
All compared methods share the student, teacher, source loss, and training
schedule, differing only in their treatment of target pixels. \emph{Source
only} omits the target loss.
\emph{All-pixel EMA} sets $w(u)=1$ for every pixel, which is standard
Mean-Teacher self-training. \emph{Original BiRP} uses the prior and teacher to
retain only pixels passing class-specific foreground and background
reliability gates and a top-$r$ selection, setting $w(u)=0$ elsewhere. The
\emph{foreground floor} variant, reported for the BUSI-source protocol, uses the original BiRP pseudo-label and blends its selected weight $m$ with a class-dependent floor $b$: $w=b+(1-b)m$, with $b=0.75$ for foreground and $b=0.25$ for background. For the
component ablation we also use \emph{certainty only}, $w=0.5+0.5c$, and
\emph{agreement only}, $w=0.5+0.5a$.

\paragraph{Training procedure.}
Each mini-batch contains labeled source images and unlabeled target images.
During warm-up, only $\mathcal L_s$ is optimized. Thereafter, the EMA teacher
and the frozen prior are evaluated on the same target images without
gradients, Eq.~\eqref{eq:weight} is computed per pixel, and the student is
updated using $\mathcal L$, after which the teacher parameters are updated by
EMA. In the main TransUNet protocol, checkpoints are selected using source-validation Dice, and evaluation uses the corresponding EMA parameters. The U-Net evaluation used for the secondary experiments is described separately in Section~\ref{sec:experiments}.

\section{Experiments}
\label{sec:experiments}
\subsection{Data and evaluation}
We evaluate on five breast ultrasound datasets: BUSI \cite{aldhabyani2020busi}, BUS-BRA \cite{gomezflores2024busbra}, BrEaST \cite{pawlowska2024breast}, OASBUD \cite{piotrzkowska2017oasbud}, and BUS-UCLM \cite{vallez2025bus}. BUSI, BUS-BRA, and BrEaST each serve as a source domain in separate experiments. Adaptation uses only unlabeled target images, with target annotations reserved for evaluation. We report the image-level mean Dice over three seeds (1337, 2027, and 2028), together with the sample standard deviation and, when available, the 95th-percentile Hausdorff distance (HD95, measured in pixels after resizing to $224\times224$). The evaluator assigns zero when both masks are empty and the image diagonal when exactly one is empty. Target cohorts differ in both partition and size. Specifically, the BUS-UCLM target comprises 128 images in the BUSI- and BrEaST-source TransUNet protocols and 126 images in the BUS-BRA-source protocol; these test cohorts are not identical. Some source partitions also vary with the target domain. We compare methods within each source--target cohort. Because the method was developed after some target-test outcomes had been inspected, all results should be interpreted as exploratory.

\paragraph{Implementation.}
\emph{TransUNet pipeline (main).} We train a TransUNet student \cite{chen2021transunet} on $224\times224$ inputs for 55 epochs, using a 10-epoch source-only warm-up, a batch size of 2, and an EMA momentum of 0.99. The controlled configuration uses AdamW with an initial learning rate of $2.5\times10^{-4}$, a weight decay of $10^{-4}$, and a target loss coefficient $\lambda_p=0.15$. For auditability, the evaluation summaries record hashes of both the checkpoints and test files. \emph{U-Net pipeline (secondary).} We train a U-Net with an ImageNet-pretrained ResNet-50 encoder \cite{ronneberger2015unet} for 8000 iterations, using 1000 warm-up iterations, a batch size of 8, AdamW with a learning rate of $2\times10^{-4}$ and weight decay of $10^{-4}$, an EMA momentum of 0.999, and a target loss coefficient $\lambda_p=1$. Both pipelines use a frozen AutoSAMUS prior \cite{lin2023autosamus}. The TransUNet experiments evaluate EMA parameters associated with the source-validation-selected checkpoint. For the U-Net results, we report the test Dice recorded for the final-iteration EMA model. Because the pipelines differ in architecture, optimization schedule, data partitions, and prior checkpoint, their results are reported separately and are not pooled.

\subsection{BUSI as source}
Table~\ref{tab:busi} compares original BiRP, all-pixel EMA, a class-dependent foreground-floor variant, and continuous agreement under the controlled BUSI-source protocol. Continuous agreement yields a higher mean Dice than both original BiRP and all-pixel EMA on all four targets, with the largest margin over all-pixel EMA observed on OASBUD (+.057). Although the foreground-floor variant achieves the highest mean on BrEaST, its performance on BUS-BRA is highly variable (.6133$\pm$.0719). These differences do not establish statistical significance, and the low absolute performance on BUS-UCLM remains an important limitation.
\begin{table}[!htbp]
\centering\footnotesize
\caption{BUSI-source TransUNet. Test Dice (mean $\pm$ sample SD, three seeds). Test sizes: 376, 51, 40, and 128 images, respectively. FG floor denotes the foreground-floor variant.}
\label{tab:busi}
\setlength{\tabcolsep}{2pt}
\begin{tabular*}{\columnwidth}{@{\extracolsep{\fill}}lcccc@{}}
\toprule
Method & BUS-BRA & BrEaST & OASBUD & BUS-UCLM \\
\midrule
Original BiRP & .7328$\pm$.0079 & .7234$\pm$.0235 & .4438$\pm$.0056 & .3333$\pm$.0119 \\
All-pixel EMA & .7736$\pm$.0129 & .7298$\pm$.0067 & .4669$\pm$.0440 & .3409$\pm$.0280 \\
FG floor & .6133$\pm$.0719 & \textbf{.7504}$\pm$.0141 & .4600$\pm$.0423 & .3741$\pm$.0169 \\
Continuous & \textbf{.7888}$\pm$.0120 & .7444$\pm$.0182 & \textbf{.5239}$\pm$.0394 & \textbf{.3768}$\pm$.0033 \\
\bottomrule
\end{tabular*}
\end{table}

\subsection{BUS-BRA as source}
Table~\ref{tab:busbra} reports four reverse adaptation directions evaluated with the same TransUNet pipeline. Among the methods compared, continuous agreement attains the highest mean Dice and the lowest mean HD95 on every target (Table~\ref{tab:hd95}). Original BiRP underperforms source-only training on BUSI, while all-pixel EMA underperforms it on BUS-UCLM; continuous agreement yields a higher mean than source-only training in all four directions. Nevertheless, performance on BUS-UCLM remains low in absolute terms, and the OASBUD estimates vary substantially across seeds. The observed rankings should therefore not be interpreted as evidence of statistical significance.
\begin{table}[!htbp]
\centering\footnotesize
\caption{BUS-BRA-source TransUNet. Test Dice (mean $\pm$ sample SD, three seeds). Test sizes: 51, 129, 40, and 126 images, respectively.}
\label{tab:busbra}
\setlength{\tabcolsep}{2pt}
\begin{tabular*}{\columnwidth}{@{\extracolsep{\fill}}lcccc@{}}
\toprule
Method & BrEaST & BUSI & OASBUD & BUS-UCLM \\
\midrule
Source only & .6980$\pm$.0167 & .6875$\pm$.0166 & .5784$\pm$.0479 & .2478$\pm$.0076 \\
Original BiRP & .7215$\pm$.0072 & .6777$\pm$.0085 & .6025$\pm$.0620 & .2618$\pm$.0033 \\
All-pixel EMA & .7030$\pm$.0065 & .6998$\pm$.0195 & .6168$\pm$.0386 & .2373$\pm$.0424 \\
Continuous & \textbf{.7359}$\pm$.0188 & \textbf{.7278}$\pm$.0107 & \textbf{.6589}$\pm$.0495 & \textbf{.2696}$\pm$.0050 \\
\bottomrule
\end{tabular*}
\end{table}

\subsection{BrEaST as source}
We further evaluate the BrEaST-source directions using the TransUNet pipeline (Tables~\ref{tab:breast} and~\ref{tab:hd95}). On BUSI, continuous agreement achieves the highest mean Dice and the lowest HD95. On OASBUD, its mean Dice is close to that of original BiRP (.6059 versus .6087), with a difference smaller than one seed-level standard deviation, and it exhibits the lowest variability across seeds. All three adaptation methods outperform source-only training on this target, for which the large standard deviation (.1000) should be interpreted in light of the 40-image test set. On BUS-UCLM, however, no adaptation method surpasses source-only training in Dice, and every adaptation method produces a higher HD95, constituting a clear negative result for this direction.
\begin{table}[!htbp]
\centering\footnotesize
\caption{BrEaST-source TransUNet. Test Dice (mean $\pm$ sample SD, three seeds). Test sizes: 126, 40, and 128 images, respectively.}
\label{tab:breast}
\setlength{\tabcolsep}{2pt}
\begin{tabular*}{\columnwidth}{@{\extracolsep{\fill}}lccc@{}}
\toprule
Method & BUSI & OASBUD & BUS-UCLM \\
\midrule
Source only & .6004$\pm$.0216 & .5412$\pm$.1000 & \textbf{.3487}$\pm$.0222 \\
Original BiRP & .6290$\pm$.0182 & \textbf{.6087}$\pm$.0392 & .3321$\pm$.0053 \\
All-pixel EMA & .6183$\pm$.0268 & .6021$\pm$.0463 & .3439$\pm$.0206 \\
Continuous & \textbf{.6360}$\pm$.0198 & .6059$\pm$.0139 & .3446$\pm$.0047 \\
\bottomrule
\end{tabular*}
\end{table}

\begin{figure*}[!t]
\centering
\newcommand{\panelheading}[1]{\makebox[.16\linewidth][c]{\strut #1}}
{\small\panelheading{Ultrasound}\panelheading{Ground truth}\panelheading{Source only}%
\panelheading{Original BiRP}\panelheading{All-pixel EMA}\panelheading{Continuous}}\par\medskip
\includegraphics[width=.96\linewidth]{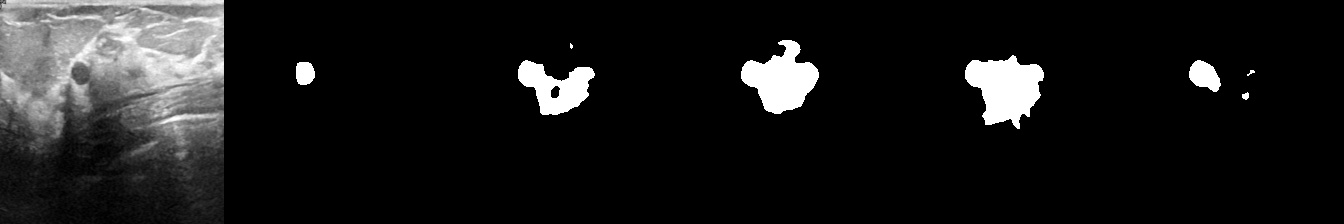}\\[2mm]
\includegraphics[width=.96\linewidth]{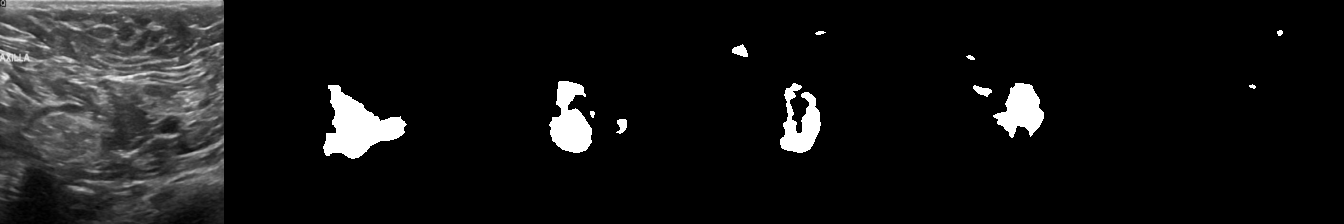}
\caption{Selected BrEaST$\rightarrow$BUSI test cases (TransUNet, seed 1337): improvement (top) and regression (bottom). Columns from left: image, ground truth, source only, original BiRP, all-pixel EMA, and continuous agreement. Cases were selected to illustrate different outcomes, not sampled to estimate their frequency.}
\label{fig:qual}\end{figure*}

\paragraph{BrEaST source in the U-Net pipeline.}
Table~\ref{tab:breast_unet} reports BrEaST-source results from the separate U-Net pipeline. Continuous agreement achieves the highest mean Dice on BUS-BRA and BUSI, although its margin over all-pixel EMA on BUS-BRA is only .0013. In contrast to the TransUNet results, all adaptation methods outperform source-only training on BUS-UCLM, where original BiRP attains the highest mean. Two single-factor controls preserve this qualitative pattern: replacing the prior checkpoint with that used in the TransUNet pipeline and training the encoder from random initialization. Thus, neither the prior checkpoint nor encoder pretraining alone reproduced the cross-pipeline discrepancy in these particular swaps. They do not rule out interactions with other settings. The U-Net BUS-UCLM test split, however, contains only 30 images, compared with 128 in the TransUNet protocol. Architecture, optimization schedule, data partitions, and test-set composition remain confounded; matched studies are needed to isolate their contributions.

\subsection{Component ablation}
We ablate the components of the weighting rule on BUSI$\rightarrow$BUS-BRA in the U-Net pipeline, using an identical, consistently ordered 370-image test split for every method and seed. As shown in Table~\ref{tab:ablation}, agreement-only weighting and the full rule yield nearly identical means (.8852 versus .8851), and both exceed all-pixel EMA by approximately .010 with low variability across seeds. Certainty-only weighting performs slightly worse than all-pixel EMA. This experiment neither isolates an additional benefit from prior certainty nor establishes statistical equivalence; instead, it suggests that agreement accounts for most of the observed effect in this setting. Original BiRP is the weakest variant, consistent with the corresponding TransUNet results. Thus, the U-Net pipeline reproduces the qualitative ranking observed for the BUSI-source TransUNet experiment in this direction, but does not constitute a controlled comparison across backbones.
\begin{table}[!htbp]\centering
\caption{Component ablation in the U-Net pipeline, BUSI$\rightarrow$BUS-BRA (370 test images), mean Dice $\pm$ sample SD over three seeds.}
\label{tab:ablation}
\begin{tabular}{llc}\toprule Method & Weight $w$ & Dice \\\midrule
Original BiRP & selective gate & .8233$\pm$.0101 \\
All-pixel EMA & $1$ & .8755$\pm$.0027 \\
Certainty only & $0.5+0.5c$ & .8739$\pm$.0017 \\
Agreement only & $0.5+0.5a$ & \textbf{.8852}$\pm$.0007 \\
Full rule & $0.5+0.5ca$ & .8851$\pm$.0043 \\\bottomrule
\end{tabular}\end{table}

\subsection{Qualitative analysis}
Figure~\ref{fig:qual} presents selected BrEaST$\rightarrow$BUSI examples from the TransUNet pipeline. In the improvement case, continuous agreement achieves Dice .669 versus .274 for the best baseline. In the regression case, continuous agreement has near-zero Dice on a malignant lesion that source-only training partially identifies (.657). Such failures illustrate a limitation of the approach, although the prediction alone does not establish the precise cause. We show the largest gain and largest loss in per-image Dice relative to the best of the three baselines for that image. These cases illustrate possible behaviors. Figure~\ref{fig:weightmap} shows the weight map for the improvement case.

\subsection{Discussion and limitations}
\label{sec:limitations}
Across the eleven TransUNet directions, continuous agreement yields a higher mean Dice than all-pixel EMA in every direction and than original BiRP in ten, with BrEaST$\rightarrow$OASBUD as the sole exception. It nevertheless fails to outperform source-only training on BrEaST$\rightarrow$BUS-UCLM; moreover, many margins are smaller than one seed-level standard deviation, and absolute Dice on BUS-UCLM remains low in the TransUNet experiments. These findings are subject to several limitations. \emph{(i) Exploratory evaluation.} The rule was developed after some target-test outcomes had been inspected, and no untouched test cohort was retained; the results therefore do not constitute independent confirmation. \emph{(ii) Statistical power.} Three seeds and small test cohorts, including only 40 OASBUD images, are insufficient to support claims of statistical significance. \emph{(iii) Baseline coverage.} Our comparisons are limited to matched self-training baselines; stronger UDA methods \cite{hoyer2022daformer,zhang2021proda} and SAM-based label-refinement approaches \cite{xu2025sammatch,zhang2024iplc} have not been evaluated under the same protocol. \emph{(iv) Pipeline dependence.} The BrEaST$\rightarrow$BUS-UCLM trend differs between the TransUNet and U-Net pipelines, and the source of this discrepancy has not been isolated. \emph{(v) Component attribution.} Agreement-only weighting matches the full rule in the single available ablation, leaving the contribution of prior certainty unresolved. \emph{(vi) Prior dependence.} Performance may depend on the selected AutoSAMUS checkpoint, and agreement between the teacher and prior can give high weight to an incorrect pseudo-label. Matched cross-backbone experiments, stronger baselines, and evaluation on independent data are therefore required before broader conclusions can be drawn.

\section{Conclusion}
We investigated continuous agreement weighting for breast ultrasound domain adaptation, in which a frozen ultrasound prior rescales the EMA teacher's pseudo-labels without discarding or altering them. Across eleven TransUNet adaptation directions derived from three source datasets, the rule achieves a higher mean Dice than matched all-pixel EMA in every direction and than original BiRP in all but one; it also reduces HD95 on every BUS-BRA-source target. These gains are modest relative to the variability across seeds, however, and the method provides no benefit over source-only training on BrEaST$\rightarrow$BUS-UCLM. Furthermore, an ablation in the separate U-Net pipeline suggests that the agreement term accounts for most of the observed effect, while the incremental contribution of prior certainty remains unclear. Because the rule was developed with access to some target-test results, the findings should be regarded as exploratory. Overall, the results motivate continuous weighting as a simple alternative to hard pseudo-label selection that incurs no additional inference cost, while underscoring the need for matched cross-backbone experiments, stronger baselines, and evaluation on independent test data.

\section*{Data and Code Availability}
The images and annotations were obtained from the original releases of BUSI, BUS-BRA, BrEaST, OASBUD, and BUS-UCLM \cite{aldhabyani2020busi,gomezflores2024busbra,pawlowska2024breast,piotrzkowska2017oasbud,vallez2025bus}. The \href{https://github.com/fanxinyusarah/birp-continuous-agreement}{GitHub repository} contains the archived TransUNet training and evaluation scripts, the recorded U-Net settings, and per-seed summaries for the reported experiments.\footnote{Generative AI tools assisted with manuscript drafting and editing and with organizing the accompanying code.} Access to the repository is currently restricted. Contact the authors for access. Obtain datasets and pretrained weights from their respective providers under the applicable terms.

{
  \small
  \bibliographystyle{ieeenat_fullname}
  \bibliography{main}
}
\onecolumn
\appendix
\section{Additional experimental results}
\label{app:additional-results}
Table~\ref{tab:hd95} reports the boundary-distance results for the BUS-BRA and BrEaST source domains. Table~\ref{tab:breast_unet} reports the separate U-Net experiments and the BUS-UCLM controls.

\begin{table}[!h]
\centering\renewcommand{\arraystretch}{1.18}
\caption{TransUNet boundary results. Mean HD95 in pixels over three seeds (lower is better).}
\label{tab:hd95}
\begin{tabular*}{0.95\linewidth}{@{\extracolsep{\fill}}lccccccc@{}}
\toprule
& \multicolumn{4}{c}{BUS-BRA source} & \multicolumn{3}{c}{BrEaST source} \\
\cmidrule(lr){2-5}\cmidrule(lr){6-8}
Method & BrEaST & BUSI & OASBUD & BUS-UCLM & BUSI & OASBUD & BUS-UCLM \\
\midrule
Source only & 41.1 & 39.0 & 45.3 & 208.4 & 45.2 & 47.6 & \textbf{172.7} \\
Original BiRP & 36.3 & 38.7 & 40.2 & 206.1 & 40.8 & 34.5 & 181.8 \\
All-pixel EMA & 41.0 & 38.4 & 37.6 & 211.5 & 42.0 & \textbf{30.2} & 178.0 \\
Continuous agreement & \textbf{33.5} & \textbf{31.1} & \textbf{32.1} & \textbf{203.4} & \textbf{38.2} & 32.3 & 178.2 \\
\bottomrule
\end{tabular*}
\end{table}

\begin{table}[!h]\centering\renewcommand{\arraystretch}{1.18}
\caption{BrEaST-source results in the separate U-Net/ResNet-50 pipeline: mean Dice $\pm$ sample standard deviation over three seeds. Test sizes: BUS-BRA 370, BUSI 129, BUS-UCLM 30. The final columns evaluate alternative prior and randomly initialized encoder controls on BUS-UCLM.}
\label{tab:breast_unet}
\begin{tabular*}{0.95\linewidth}{@{\extracolsep{\fill}}lccccc@{}}\toprule
& BUS-BRA & BUSI & BUS-UCLM & UCLM alt. prior & UCLM no pretrain \\\midrule
Source only & .8109$\pm$.0182 & .7329$\pm$.0081 & .5858$\pm$.0206 & .5942$\pm$.0308 & .5435$\pm$.0114 \\
Original BiRP & .8268$\pm$.0138 & .7652$\pm$.0153 & .6481$\pm$.0251 & .6344$\pm$.0312 & .5984$\pm$.0183 \\
All-pixel EMA & .8742$\pm$.0039 & .7682$\pm$.0010 & .6313$\pm$.0086 & .6387$\pm$.0032 & .5654$\pm$.0233 \\
Continuous & .8755$\pm$.0056 & .7770$\pm$.0011 & .6457$\pm$.0291 & .6423$\pm$.0183 & .6006$\pm$.0114 \\\bottomrule
\end{tabular*}\end{table}

\end{document}